\documentclass{article}
\usepackage{authblk}
\usepackage{amsfonts}
\usepackage{algorithmic}

\usepackage{hyperref}
\usepackage[utf8]{inputenc}
\usepackage{subcaption}
\usepackage[textwidth=14cm]{geometry}
\usepackage{float, graphicx, amsmath, xcolor, caption, subcaption, hyperref}
\usepackage[sorting=none]{biblatex}
\bibliography{references} % Entries are in the refs.bib file

\title{Differentiable Generalised Predictive Coding}
\author[1*]{Andre Ofner}
\author[1]{Sebastian Stober}
\affil{Otto-von-Guericke University, Magdeburg, Germany \protect\\ ofner@ovgu.de}

\begin{document}

\maketitle

This paper deals with differentiable dynamical models congruent with neural process theories that cast brain function as the hierarchical refinement of an internal generative model explaining observations. Our work extends existing implementations of gradient-based predictive coding with automatic differentiation and allows to integrate deep neural networks for non-linear state parameterization. Gradient-based predictive coding optimises inferred states and weights locally in for each layer by optimising precision-weighted prediction errors that propagate from stimuli towards latent states. Predictions flow backwards, from latent states towards lower layers. The model suggested here optimises hierarchical and dynamical predictions of latent states. Hierarchical predictions encode expected content and hierarchical structure. Dynamical predictions capture changes in the encoded content along with higher order derivatives. Hierarchical and dynamical predictions interact and address different aspects of the same latent states. We apply the model to various perception and planning tasks on sequential data and show their mutual dependence. In particular, we demonstrate how learning sampling distances in parallel address meaningful locations data sampled at discrete time steps. We discuss possibilities to relax the assumption of linear hierarchies in favor of more flexible graph structure with emergent properties. We compare the granular structure of the model with canonical microcircuits describing predictive coding in biological networks and review the connection to Markov Blankets as a tool to characterize modularity. A final section sketches out ideas for efficient perception and planning in nested spatio-temporal hierarchies.
We open source the Torch code for the suggested generalized predictive coding optimizer, GPC and hope that our work initiates more research on generalized state-space models in the context of deep neural networks, reinforcement learning and aspects of biological plausibility.

\section{Introduction}

\subsection{Generalized Filtering and Dynamical Generative Models}

The hierarchical dynamical generative models (HDMs) discussed by Friston and Kiebel \cite{friston2008hierarchical} in the context of explaining brain function describe a powerful class of Bayesian generative models that are hierarchical in structure and dynamics and allow to perform Bayesian inference on complex observations. HDMs capture latent states, encoding estimated causes for observations and their associated motion in generalised coordinates, i.e. the derivatives of each encoded state's trajectory. In the context of HDMS, \emph{perception} is referred to as the process of inverting a particular model and refining its parameters, in order to more accurately explain the observations. As we will see later in this article, planning and \emph{actions} are closely intertwined with perception when applying HDMs to sensory data prediction.

Interestingly, HDMs generalize most established statistical models, e.g. (extended) Kalman filters can be interpreted as a particular first order HDM. More generally speaking, it is possible to characterize most existing Bayesian model optimisation schemes as a variant of \emph{Generalized Filtering} \cite{friston2010generalised}. Generalised filtering describes a generic approach to compute posterior densities for latent states and model parameters (i.e. addressing inference and learning) based on a gradient descent on the variational Free Energy (also known as the Evidence Lower bound or ELBO in the machine learning domain). 

Without any additional assumptions about the underlying probability distributions, the variational Free Energy can be expressed as:

\begin{equation}
\begin{aligned}
\mathcal{F} &=D_{\mathrm{KL}}[q(x \mid o ; \phi) \| p(o, x ; \theta)]
&=\underbrace{\mathbb{E}_{q(x \mid o ; \phi)}[\ln q(x \mid o ; \phi)]}_{\text {Entropy }}-\underbrace{\mathbb{E}_{q(x \mid o ; \phi)}[\ln p(o, x ; \theta)]}_{\text {Energy }}
\end{aligned}
\end{equation}

with data $o$, latent state $x$ and a generative model that generates observed data $p(o, x)=p(o \mid x) p(x)$. In variational inference, the true psosterior of latent state x $p(x \mid o)=\frac{p(o, x)}{p(o)}$ is inferred indirectly using an approximate posterior $q(x \mid o ; \phi)$, with known parameters $\phi$. To do so, fitting the approximate posterior to the true posterior is achieved by minimizing the divergence between the true and approximate posterior with respect to the parameters 

\begin{equation}
D_{\mathrm{KL}}(q(x \mid o ; \phi) \| p(o, x ; \theta))
\end{equation}

Existing models that implement Generalized Gradient Descent on the Free Energy take different forms and with that, different degrees of biological plausibility - at different levels of analysis. 
Important variants implementing generalized filtering are variational filtering \cite{friston2008variational}, dynamic expectation maximization \cite{friston2008variational} and generalized predictive coding \cite{friston2009predictive}. Of these, generalized (Bayesian) predictive coding is a particularly interesting candidate, since it allows relatively straightforward mapping to neural mechanism. 

We will focus on generalized predictive coding in this article, but keep the similarities to related, more generic filtering schemes in mind. Finally, HDMs typically rest on functions operating in continuous time, which requires additional mechanisms for updates with respect to discrete time intervals, such as they are present in many machine learning problems of the kind that we want to look at here. 

Dynamical generative models $p(y, \vartheta)=p(y \mid \vartheta) p(\vartheta)$ describe the likelihood $p(y \mid \vartheta)$ of observing data $y$ given causes $\vartheta=\{x, v,\theta\}$, and priors on causes $p(\vartheta)$. With non-linear functions $h$ and $d$, parameterized $\theta_h$ and $\theta_d$, these models generate outputs (responses) $\widetilde{y}=\left[y, y^{\prime}, y^{\prime \prime}, \ldots\right]$ characterised by the first derivatives of their trajectory, the generalized coordinates of motion:

\begin{equation}
\begin{aligned}
y &= h(x, v)+z\\
y^{\prime} &= h_{x} x^{\prime}+h_{v} v^{\prime}+z^{\prime} \\
y^{\prime \prime} &= h_{x} x^{\prime \prime}+h_{v} v^{\prime \prime}+z^{\prime \prime}\\
\vdots
\end{aligned}
\end{equation}

with stochastic observation noise z. Similarly, the generalized coordinates of the motion of cause states are

\begin{equation}
\begin{aligned}
x^{\prime} &= d(x, v)+w \\
x^{\prime \prime} &=d_{x} x^{\prime}+d_{v} v^{\prime}+w^{\prime} \\
x^{\prime \prime \prime} &= d_{x} x^{\prime \prime}+d_{v} v^{\prime \prime}+w^{\prime \prime} \\
\vdots
\end{aligned}
\end{equation}

with stochastic transition noise w. Here, the cause states $\vartheta=\{x, v, \theta\}$ split into input states $v(t)$ at time $t$ and hidden states $x(t)$ that are part of a transition function $d$ that projects input states to output states. From a machine learning perspective, we can directly associate $x(t)$ with the hidden states or memory in Recurrent Neural Networks (RNNs) and $d(x)$ with the corresponding transition function in RNNs. Differently to the single-order transition function in classical RNNs, the function $d$ couples multiple orders of dynamics.

Next to a hierarchy of the cause state dynamics, the models described in \cite{friston2009predictive} model a hierarchical structure of the cause states themselves:

\begin{equation}
\begin{array}{c}
v^{i-l}=g\left(x^{l}, v^{l}\right)+z^{l} \\
\dot{x}^{l}=f\left(x^{l}, v^{l}\right)+w^{l}
\end{array}
\end{equation}

where input and hidden states $v^l$ and $x^l$ at layer $l$ link states and dynamics between layers $l$. In such hierarchical models, the outputs of a layer are the inputs to the next lower layer and the fluctuations $z^l$ and $w^l$ influence the fluctuations of states at the next higher level. 

\subsection{Inverting hierarchical dynamical models with Generalized Predictive Coding}

The described general structure of hierarchical dynamical models does not yet entail the specific process that actually learns parameters and states inside the model from observations. This process, called model inversion, can be described at different levels of description, each of which leads to different assumptions at the implementations level. A general approach for model inversion is based on variational Bayesian inference, where the conditional density $p(\vartheta \mid y, m)$ for causes $\vartheta$, model $m$ and data $y$ is approximated using a recognition density $q(\vartheta)$ with respect to a lower bound on the evidence $p(y \mid m)$ of the model. More detailed descriptions of hierarchical model inversion can be found, for example, in \cite{friston2008hierarchical}.
When this Bayesian inversion is done with respect to (explicitly represented) prediction errors, this process can be implemented with predictive coding. (Bayesian) predictive coding is a algorithmic motif that is plausibly represented in biological neurons and thus provides an interesting candidate for improving machine learning systems. 

Generalized predictive coding can be described as a gradient descent on precision weighted prediction errors \cite{friston2008hierarchical, bastos2012canonical}:

\begin{equation}
\begin{aligned}
\dot{\tilde{\mu}}_{v}^{l}&=\tilde{\mu}_{v}^{l}-\partial_{\widetilde{v}} \widetilde{\varepsilon}^{l} \cdot \xi^{l}-\xi_{v}^{(l+1)}\\
\dot{\tilde{\mu}}_{x}^{l}&=\tilde{\mu}_{x}^{l}-\partial_{\tilde{x}} \widetilde{\varepsilon}^{l} \cdot \xi^{l}\\
\widetilde{\varepsilon}_{v}^{l}&=\widetilde{\mu}_{v}^{(l-1)}-g^{l}\left(\widetilde{\mu}_{x}^{l}, \widetilde{\mu}_{v}^{l}\right) \\
\widetilde{\varepsilon}_{x}^{l}&=\tilde{\mu}_{x}^{l}-f^{l}\left(\widetilde{\mu}_{x}^{l}, \widetilde{\mu}_{v}^{l}\right)\\
\xi_{v}^{l}&=\Sigma^{-1}_{{v}^{l}} \widetilde{\varepsilon}_{v}^{l} \\
\xi_{x}^{l}&=\Sigma^{-1}_{{x}^{l}} \widetilde{\varepsilon}_{x}^{l}\\
\end{aligned}
\end{equation}

where $l$ is the current layer and $\Sigma^{-1}$ is the inverse variance (the precision) of the prediction errors $\widetilde{\varepsilon}$ in generalized coordinates. The precision weighted prediction errors $\xi_{v}$ and $\xi_{x}$ are generated with respect to the difference between current expectations about hidden causes and states $\tilde{\mu}_{v}$ and $\tilde{\mu}_{x}$ and their predicted values. Predicted values of causes and states are computed with non-linear feedback functions $f$ and $g$. Crucially, the dynamics of the hidden states in a particular layer are computed with respect to the states in the same layer. In contrast, cause states are updated with respect to predictions that map from states of the next higher layer. Effectively, this means that hidden states model the dynamics within layers, while cause states link different layers. \footnote{This restriction to roles in intrinsic and extrinsic connectivity for cause and hidden states comes from assumptions about the underlying graphical model, where the dynamics of different nodes can strictly be separated. In biological and artificial implementations, these boundaries might be much looser and under constant change. As we'll see in the discussion of this article, the interpretation of neural (cortical) function in terms of canonical predictive coding microcircuits might be substantially less straightforward than it appears from the math \cite{bastos2012canonical}}.

Existing models that implement a particular form of generalized filtering need to make specific assumptions, particularly about the nature of modelled functions and the computation of their generalized coordinates. A popular approach is to resort to numerical differentiation with the finite difference method. Numerical differentiation, however, often struggles with rounding and discretisation problems and the exact computation of higher order derivatives in general. Automatic differentiation, which drives a substantial amount of state-of-the art machine learning solves these problems by operating on functions with known exact gradients. It is this technique that we want to adapt here in the context of HDMs.

While several implementations of generalized filtering variants exist, they do not quite yet scale to complex applications such as perception of complex data, or reinforcement learning problems such as dealt with in the deep learning domain. Here, we want to focus on implementing hierarchical dynamical models with update rules described by gradient based predictive coding. Specifically, we look at Stochastic Gradient Descent (SGD) on prediction errors with exact gradients in differentiable deep neural network (DNN) models.

\subsection{Differentiable predictive coding and deep neural networks}

Much effort has recently been spent on developing scaled up versions of gradient based predictive coding networks and comparing their (relatively) biologically plausible updates with the exact gradient computation involved in the backpropagation algorithm that drives a substantial amount of current state-of-the art machine learning \cite{millidge2021predictivereview, rosenbaum2021relationship, bogacz2017tutorial, ofner2021predprop, rumelhart1986learning}. In particular, it has been shown that, under specific assumptions \footnote{These assumptions include fixing predictions during weights updates or the inversion of data and target inputs to the model - so called "discriminative" predictive coding.} the weight updates in predictive coding networks use approximations of the exact gradients in backpropagation \cite{millidge2020predictive, rosenbaum2021relationship, whittington2017approximation}. Similar experiments for the gradients in HDMs have been, so far, not been conducted. Related efforts have resulted in optimization strategies that allow to turn (deep) neural networks into gradient based predictive coding networks \cite{millidge2020predictive, rosenbaum2021relationship}. PredProp, a DNN optimiser proposed in \cite{ofner2021predprop} jointly optimizes each layer's learnable parameters in parallel using exact gradients. This approach removes the need to wait for individual variables to converge (such as the fixed prediction assumption when training weight parameters) and allows for constant interaction of the estimated precision errors between layers. Another aspect of PredProp is the possibility to include complex DNN architectures within the backward prediction weights of a single predictive layer, while layer-wise predictive coding \footnote{Which is arguably more biologically plausible due to the significantly reduced computational complexity in the connections between state variables.} is still an option. The model presented here is an extension to the PredProp optimiser that focuses on adding a dynamical pathway in models with simple feedback and transition weights. 
Outside of gradient based predictive coding, there are various models that include predictive coding inspired mechanisms into state-of-the-art deep neural architectures, although many times with substantial differences to the mechanisms described in the predictive coding theory \cite{lotter2016deep, millidge2021predictivereview, rane2020prednet}. Detailed reviews on predictive coding variants and their connection to established methods in machine learning, such as variational autoencoders and normalizing flows can be found in \cite{millidge2021predictivereview, marino2020predictive}. 

\section{Inferring content and dynamics with generalised predictive coding }
\begin{figure}
\begin{centering}
{a)}\includegraphics[width=0.7\textwidth]{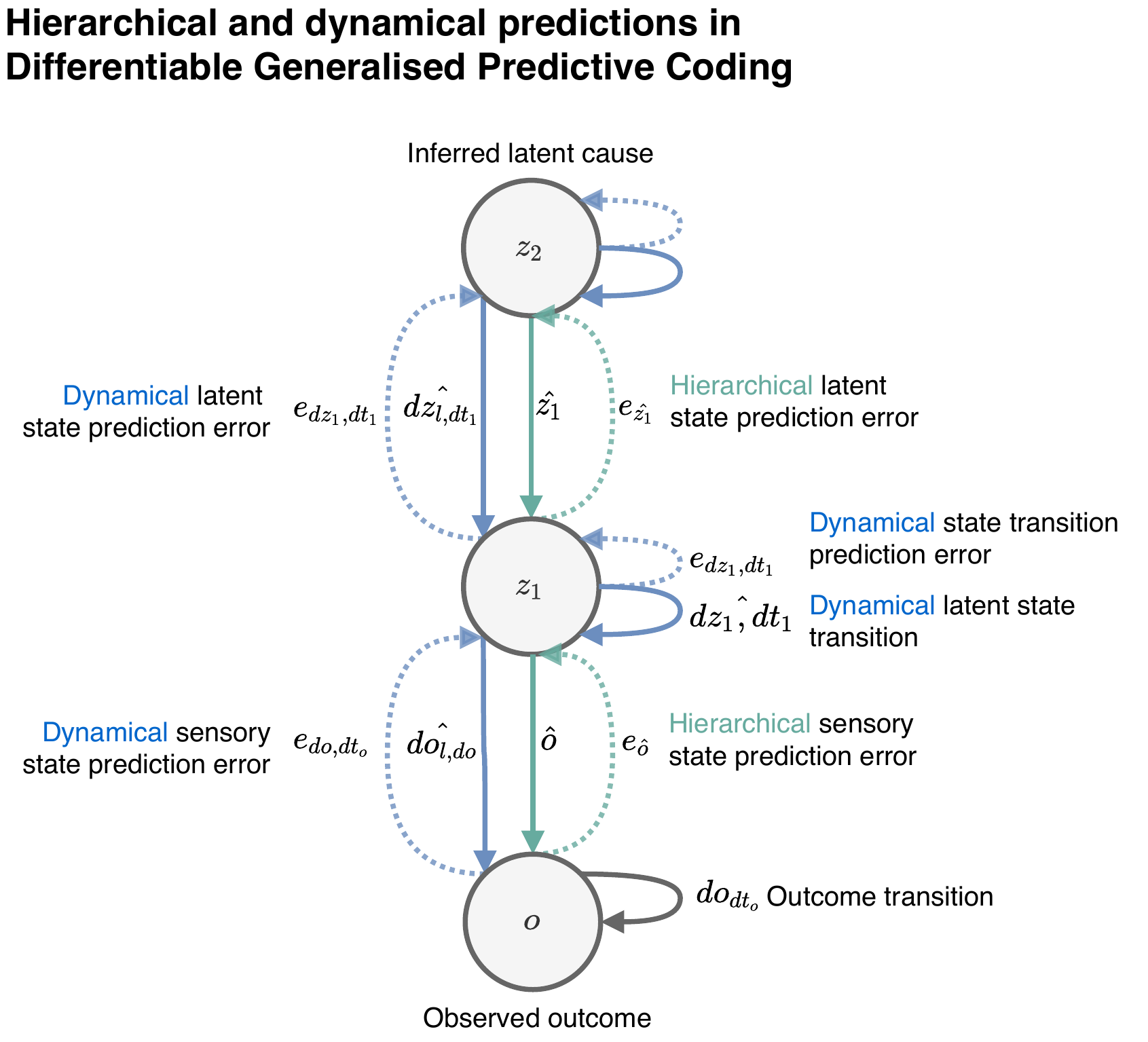}
{b)}\includegraphics[width=0.6\textwidth]{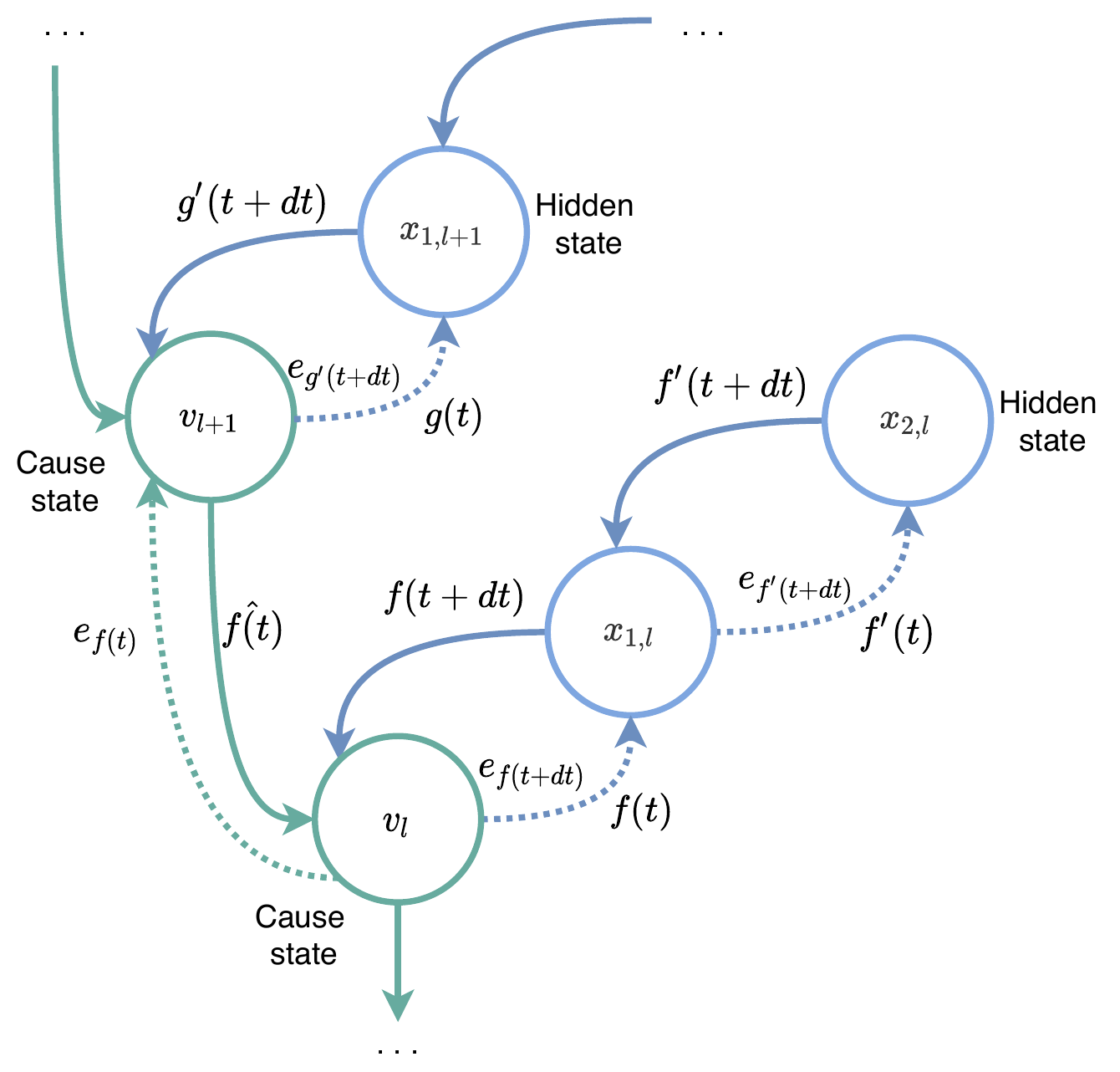}
\caption{a) Hierarchical and dynamical predictions in Differentiable GPC. Blue and green colors indicate dynamical and hierarchical pathways respectively. Dotted arrows indicate precision-weighted prediction errors that drive local updates of learnable parameters. Here, cause states and hidden states $v_l$ and $x_l$ are summarized as states $z_l = [v_l, x_l]$ in \emph{layer} l. b) Different \emph{levels} of causes can be decoupled from their dynamics, when higher layers predict inputs (green), but do not predict their dynamics (blue). For example, the motion ($x_l$) of birds ($v_l$) flying might be associated with the relatively static concept of nature ($v_{l+1}$) as a cause.}
\label{fig:diagram}
\end{centering}
\end{figure}

We introduce an optimisation method based on Stochastic Gradient Descent, called GPC, that implements Generalised Predictive Coding with exact gradients along the hierarchical and dynamical dimension, while still offering the possibility to estimate the true dynamics underlying raw sensory data. GPC augments an existing predictive coding based optimisation method called PredProp \cite{ofner2021predprop} and employs exact gradients to optimize inferred states, weights and prediction error precision in a layer-wise fashion and in parallel (i.e. without having to wait for other variables to converge) for arbitrary neural network architectures. Figure  
\ref{fig:diagram} summarizes the main components of the proposed optimisation scheme in a predictive coding model with two layers.

For the sake of completeness, we summarize the core functions of PredProp here. For a more detailed description of PredProp, the reader is referred to \cite{ofner2021predprop}. PredProp implements stochastic gradient descent on the negative Free Energy. Prediction errors $\epsilon_{l}$ at each layer, weighted by their \emph{precision} (inverse covariance) $\Sigma_{l}^{-1}$ are minimized for all layers \emph{individually}.

More precisely, under the Laplace assumption, the variational Free Energy simplifies to a Free Energy term based on prediction errors, which can be noted as:

\begin{equation}
\mathcal{F}=-\sum_{l=1}^{L} \Sigma_{l}^{-1} \epsilon_{l}^{2}+\ln 2 \pi \Sigma_{l}^{-1}
\end{equation}

with prediction errors $\epsilon_{l}$ between an observed activity $\mu_{l}$ and predicted mean activity $\hat{\mu}_{l}=f_{l}(\theta_{l+1}, \mu_{l+1})$ at layer $l$:
\begin{equation}
\epsilon_{l}=\mu_{l}-f_{l}\left(\theta_{l+1}, \mu_{l+1}\right)
\end{equation}
More details on the variational Free Energy and the Laplace assumption in the context of predictive coding are discussed in \cite{friston2006free, millidge2021predictivereview, bogacz2017tutorial}.

Effectively, a cost function is optimized for each layer in parallel. As predictive coding are linked via shared cause states, the summed cost function of the entire network is optimized simultaneously. PredProp optimises networks bidirectionally, so propagated values constantly iterate between inferring the most likely cause (i.e.~target) of observed data and the most likely data given the currently estimated cause. Each predictive coding layer $l$ has at least one or multiple weights parameterizing the backwards connection $\theta$, followed by a (non-)linear activation function $f$. Theoretically, these multi-layer backward connections can incorporate entire DNN structures, since exact gradients can be computed with automatic differentiation. 

In a single-layer predictive coding network, the observed activity $\mu$ simply is the input data. Layers in multi-layer predictive coding networks predict the activity $\mu_{l-1}$ of the next lower layer. The backward predictions of the next lower layer are computed using a (non-)linear activation function $f_l$ and learnable weights $\theta_{l}$. The resulting update at time $t$ for the backward weights $\theta_{l}$ at layer $l$ is

\begin{equation}
\begin{aligned}
\frac{d \theta_{l+1}}{d t} =\frac{\partial \mathcal{F}}{\partial \theta_{l}}=-\Sigma_{l-1}^{-1} \epsilon_{{l-1}} \frac{\partial f}{\partial \theta_{l}} \mu_{l}^{T}
\end{aligned}
\end{equation}

Next to updating the weights (i.e. parameter learning), PredProp performs a simultaneous Gradient Descent on activity (i.e. state inference) and precision in each layer directly. The corresponding update rules for activity $\mu_l$ and precision $\Sigma_{l}^{-1}$ at layer $l$ are:
\begin{equation}
\begin{aligned}
\frac{d \mu}{d t} &=-\frac{\partial \mathcal{F}}{\partial \mu}=\Sigma_{l}^{-1} \epsilon_{l} \frac{\partial f}{\partial \mu_{l+1}} \theta_{l+1}^{T}-\Sigma_{l}^{-1} \epsilon_{l} \\
\frac{d \Sigma}{d t}&=-\frac{\partial \mathcal{F}}{\partial \Sigma} = {\Sigma_{l}^{-1} \epsilon}^{2}- \Sigma_{l}
\end{aligned}
\end{equation}

As discussed in more detail in \cite{millidge2021predictivereview, ofner2021predictive}, the estimated precision of the prediction error plays an important role in predictive coding, since it estimates second-order changes in the partial objective functions and weights parameter updates accordingly.

\subsection{From static to dynamic predictive coding models}

\begin{figure}
\begin{centering}
\includegraphics[width=0.7\textwidth]{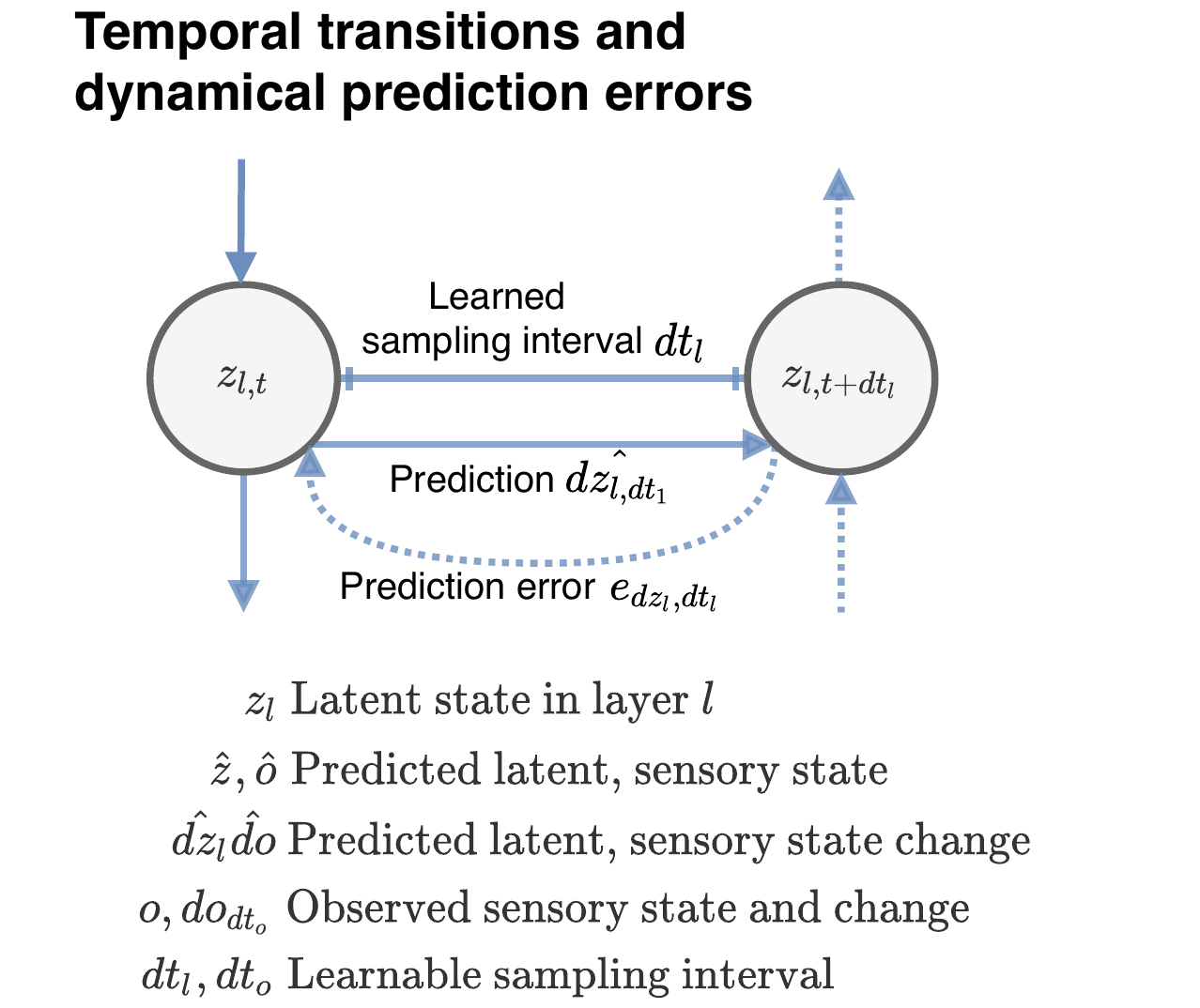}
\caption{Temporal transitions and associated dynamical prediction errors occurring in the transition function $d_l$ at each dynamical layer $l$ of the predictive coding network.}
\label{fig:transition_diagram}
\end{centering}
\end{figure}

In contrast to PredProp, which so far has been applied to static perception tasks, GPC addresses changes in observed inputs. Like many deep neural architectures, GPC operates on discrete time steps. However, the transitions inside the model have adaptive interval sizes, which are modulated only by the respective layer individually, in order to improve its own predictions as well as its top-down predictability. Since deeper dynamical layers in GPC encode \emph{known} analytic functions via their feedback weights, discrete changes over such adaptive intervals can efficiently be encoded. Importantly, at each hierarchical layer, multiple instance operate in parallel and randomly sample new intervals modulated by the precision of prediction errors. Figure \ref{fig:transition_diagram} shows the learnable variables that are relevant to the transition function at an individual predictive coding layer.  

Each layer $l$ estimates a state $x$ at discrete time $t$ for time increments $dt$ of arbitrary size. Additionally, each layer learns transition weights $W_{d}$ that paramterize the state transition function $d$

\begin{equation}
x_{l,t} = d(x_{l,t-dt}, dt) + s
\end{equation}

where $u$ is the stochastic noise driving the selection of sampling intervals (the step size or stride in the context of DNN models with discrete timesteps).

Weights $W_{h}$ parameterize the hierarchical prediction function $h$:

\begin{equation}
x_{l,t} = h(x_{l+1,t}) = W_h(x_{l+1,t})
\end{equation}

Functions $d,h$ parameterize a single connection along the dynamical and hierarchical pathway respectively. Layers along the hierarchical dimension are connected as each layer's hierarchical prediction is the expected state at thse next lower layer $x_{l,t}$. Neighboring layers along the dynamical dimension are connected as each layer predicts the derivative of the transition function of the next lower layer $d(x_{l,t})$. 
The discrete time increment $dt$ plays an important role, since it influences the estimated derivative of the observed activity, by paying attention to changes across different temporal scales. 

This estimated derivative is approximated by transition weights $\theta_{t}$. Since the learned transition function is known (in contrast to the function generating the observed data), so is its exact gradient. In the dynamical pathway, neighboring layers predict this gradient of the next lower layer. Alternatively to computing the gradient, it is possible to simply predict the change that the transition function applies to known state inputs:

\begin{equation}
\begin{aligned}
d'(x_{l,t}) &= \frac{d(x_{l,t}, dt) - d(x_{l,t-dt}, dt)}{dt}\\
&= \frac{d(x_{l,t}, dt) - x_{l,t}}{dt}\\
\end{aligned}
\end{equation}.

Here, we use that $x_{l,t} = W_{d} x_{l,t-dt}$, i.e. we require the transition weights to be estimated with respect to a known time increment $dt$. Initially $dt$ is selected randomly based on each layer's prior on the sampling interval fluctuation $s$. Since multiple instances of a layer can predict the same input, different sampling intervals $dt$ are covered at any point in time. The interval $dt$ is known in the local context of a layer, but is not propagated to distant layers. As each layer's states are predicted top-down, the optimal inferred sampling interval $dt$ balances a low prediction error for the outgoing prediction with the incoming top-down prediction.

Dynamical and hierarchical pathways are mutually interacting:

\begin{equation}
\begin{aligned}
x_{l,t} &= d(x_{l,t-dt}, dt) = d(h(x_{l+1,t-dt}), dt)\\
&= h(x_{l+1,t}) =  h(d(x_{l+1,t}, dt))\\
\end{aligned}
\end{equation}

At discrete time step $t$, the total prediction error in each layer is the sum of dynamical and hierarchical prediction errors:

\begin{equation}
\begin{aligned}
e_{l} &= e_{x,l} + e_{d,l} + e_{d',l}\\
&= (x-h(x_{l+1,t})) + (x-d(x_{l,t-dt}) + (d'(x_{l,t})-\hat{d'(x_{l,t}}))\\
&= (x-\hat{x}_h) + (x-\hat{x}_d) + (dx-\hat{dx_d})\\
\end{aligned}
\end{equation}

where $\hat{x}_h, \hat{x}_d $ and $\hat{dx_d}$ refer to the hierarchically predicted state, the transitioned state and the predicted state change of the last state transition respectively.

Given this objective function for each predictive coding layer, we can apply the optimised network to various learning and inference tasks. Optimising multiple sample intervals in parallel is thought to reduce the need for more complex autoregressive modelling, by trading computational complexity within single layers with increased parallelism. To test this hypothesis, a first step is to apply the network to the prediction of sequential data. It should be noted that, although we focus on sampling temporal inputs, the same method can be applied to spatial dimensions. The following section evaluate dynamics learning from sequential data, with and without sensory noise and non-linear activation functions.

\section{Learning dynamics from sequential data}

\subsection{Learning dynamics with linear and non-linear activation}

\begin{figure}
\includegraphics[width=\textwidth]{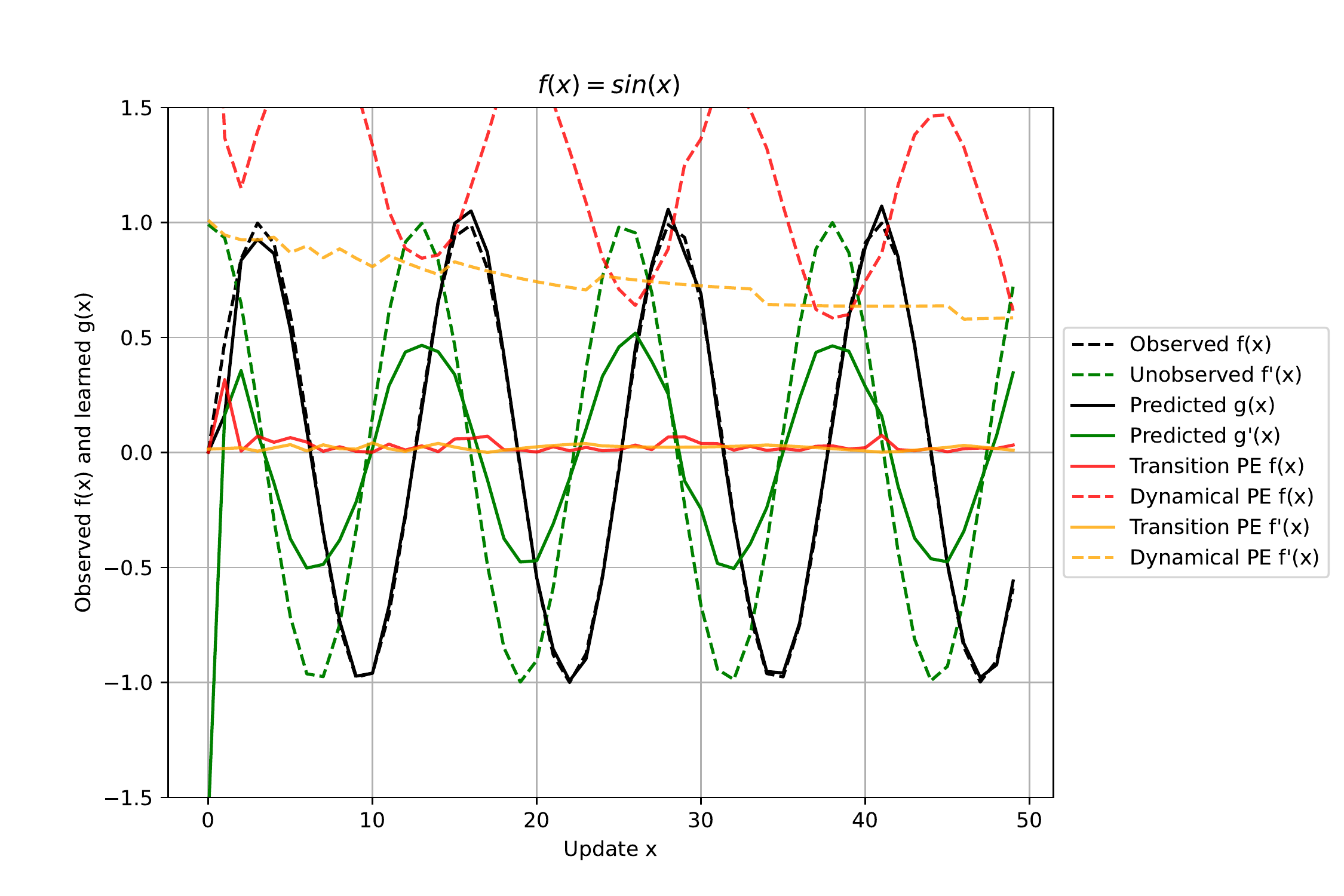}
\includegraphics[width=\textwidth]{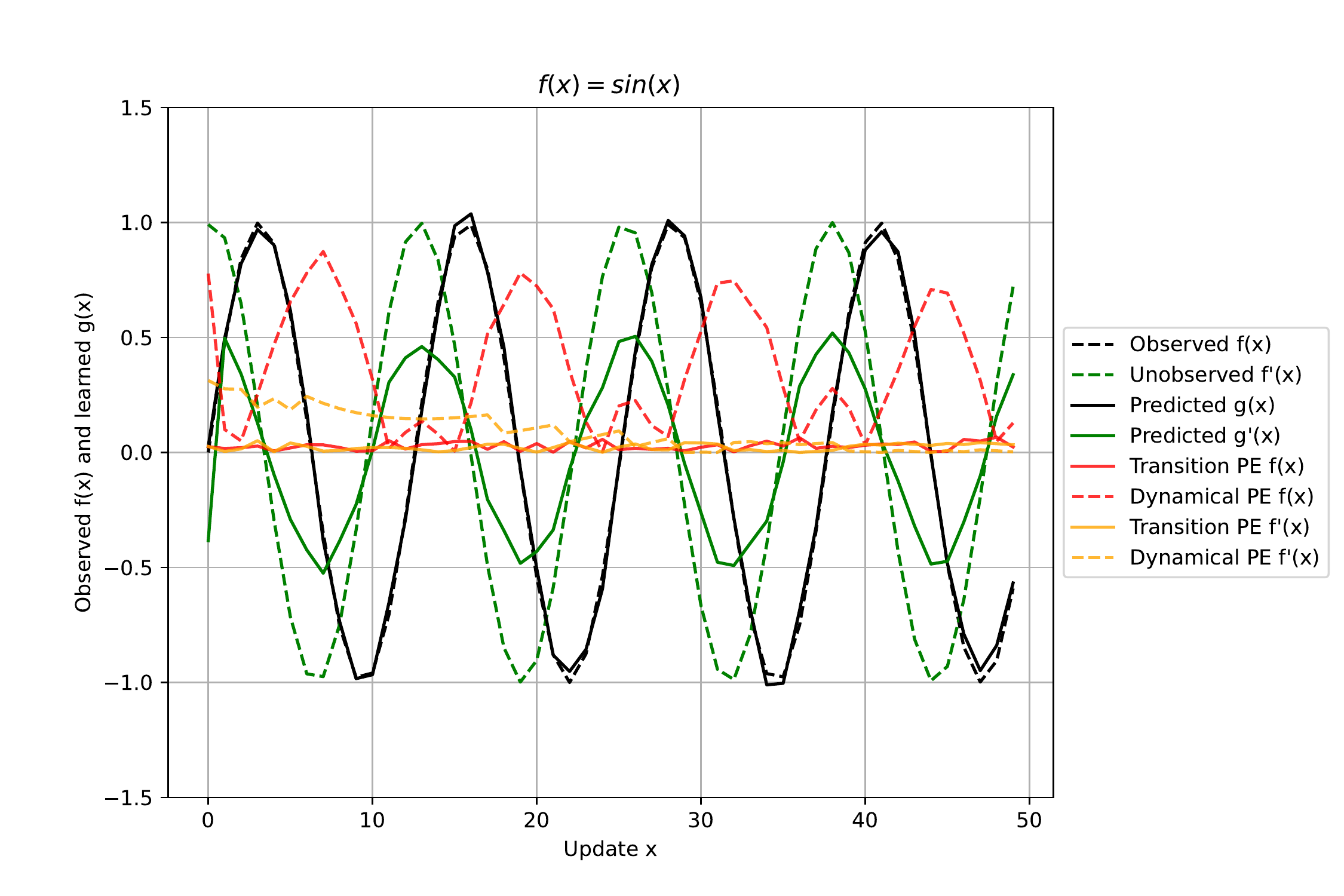}
\caption{Approximated representation of a sine function in a GPC model with three hidden dynamical layers (black and green lines). The feedback predictions are linearly activated (top) or ReLU activated (bottom). Prediction errors from the transition functions $d_1$ and $d'_2$ as well as their top-down predictions $\hat{d_2}$ and $\hat{d'_2}$ are indicated in red and yellow respectively.}
\label{fig:deriv_learning}
\end{figure}

With linear activations, GPC infers observed $f(x)$ and the first two derivatives $f'(x), f''(x)$ relatively exactly. When a ReLU non-linearity is used, the represented function has less complex generalised coordinates, but still matches the observed $f(x)$ well. Figure \ref{fig:deriv_learning} show examples for dynamical state estimation on sequential inputs with and without non-linear activation functions.

\subsection{Learning dynamics with parallel sampling}

\begin{figure}
\includegraphics[width=\textwidth]{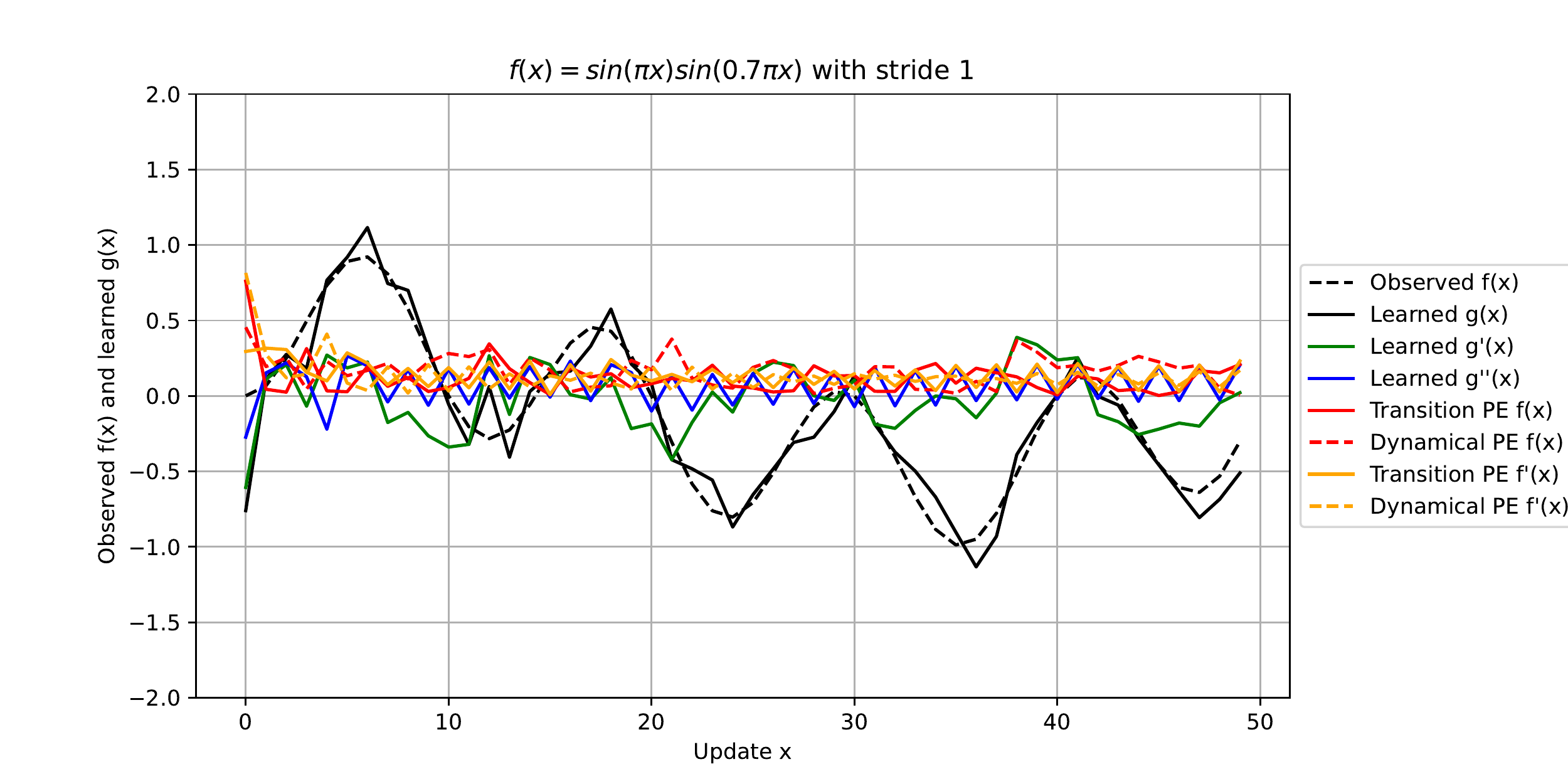}
\includegraphics[width=\textwidth]{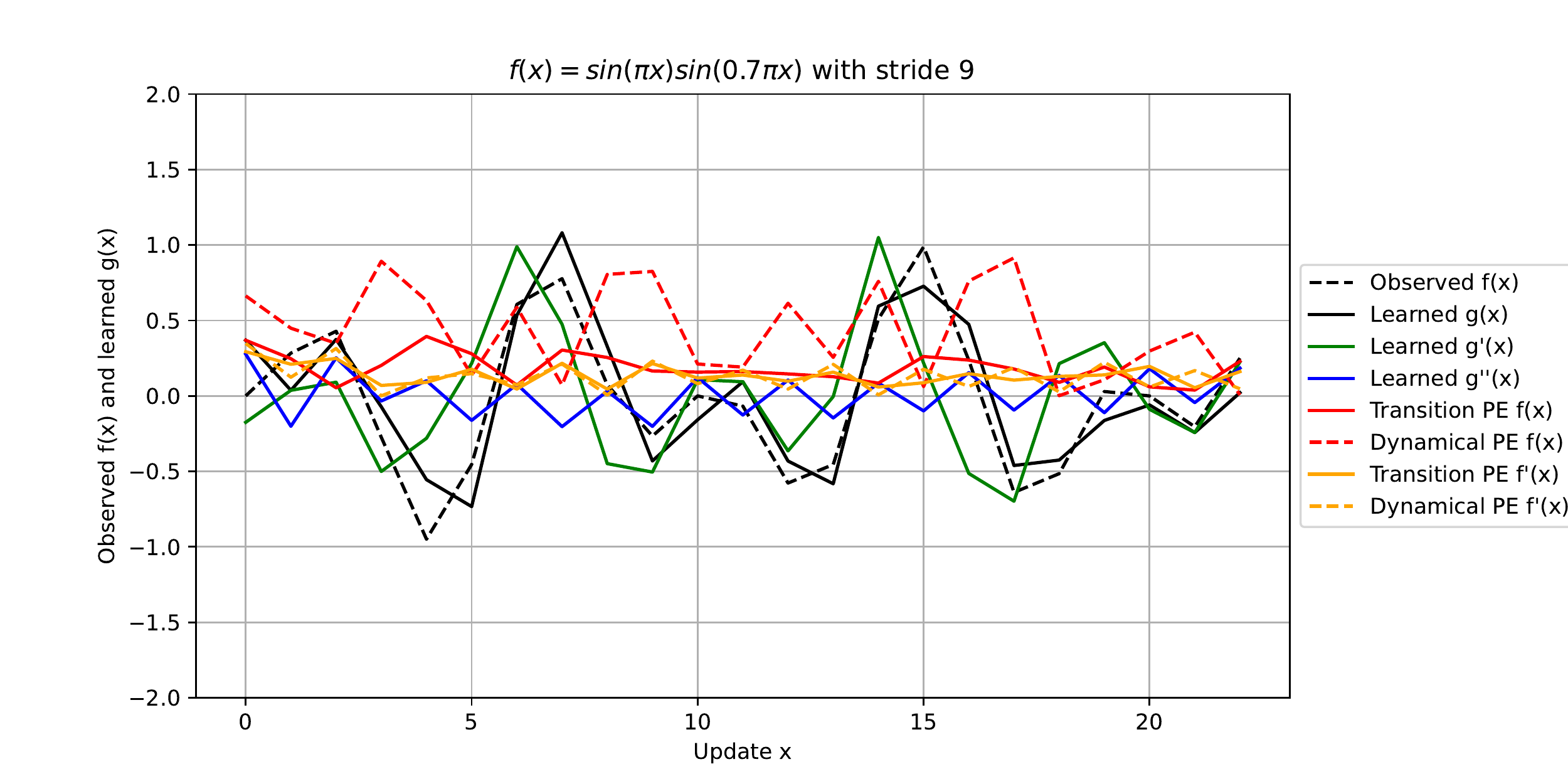}
\includegraphics[width=\textwidth]{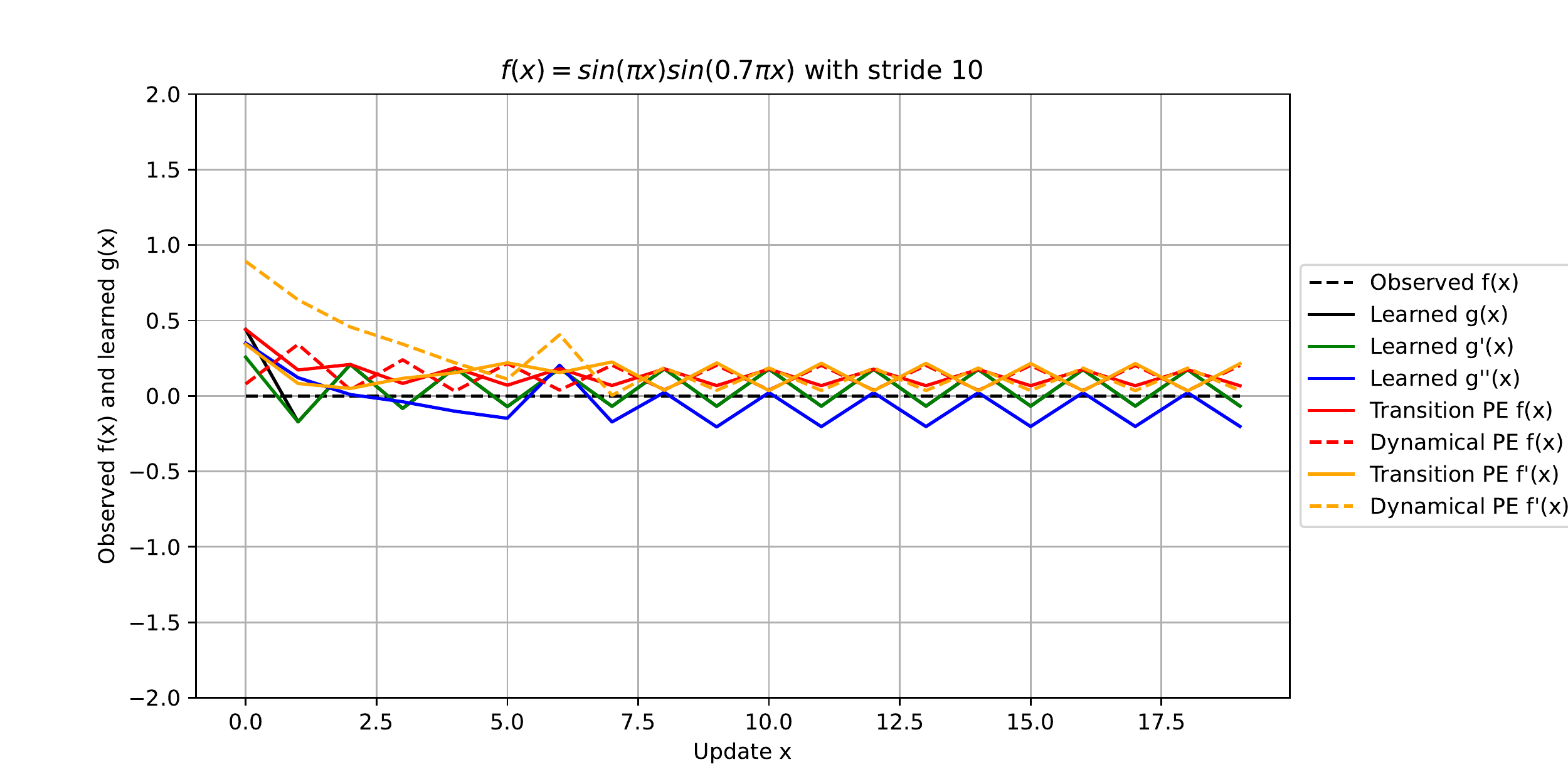}
\caption{ Dynamical state estimation on a modulated sine wave sampled at three different strides (1,9,10) in the lowest predictive coding layer. }
\label{fig:dynamics_learning}
\end{figure}

With changing interval size (stride), the inferred generalised coordinates change significantly. In the example shown in Figure \ref{fig:dynamics_learning}, a stride of 10 leads to a relatively constant stationary input, where the second represented derivative remains nearly constant. For strides other than 5 or 10, the prediction error is substantially higher.

%\subsection{Learning dynamics with adaptive sampling}

\section{Relation to the canonical microcircuit for predictive coding in biological networks}

\begin{figure}[H]
\includegraphics[width=\textwidth]{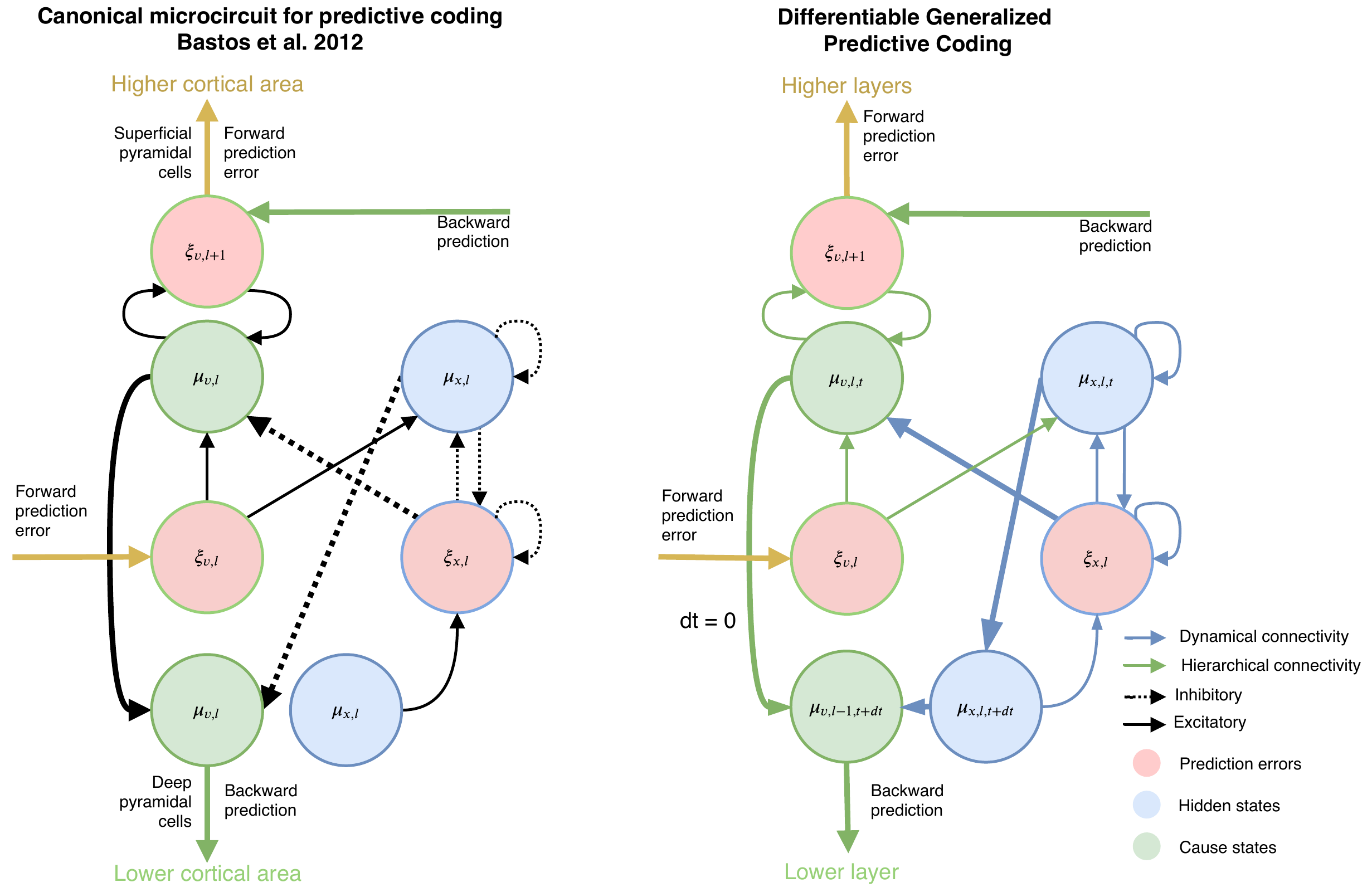}
\caption{ Comparison of Differentiable Generalized Predictive Coding with the canonical microcircuit for predictive coding that is hypothesized to drive Bayesian inference in the human brain.}
\label{fig:microcircuit}
\end{figure}

A lot of effort has been spent on describing canonical microcircuits that capture the core functionality of neuronal processing in human brains. In the context of interpreting neuronal computations as performing Bayesian inference, multiple similar models of canonical computations have been suggested that show close correspondence to the connectivity described by generalised predictive coding \cite{douglas1991functional, douglas2004neuronal, douglas1989canonical, bastos2012canonical}. 

Figure \ref{fig:microcircuit} shows a comparison between GPC, the Differentiable Generalised Predictive Coding model suggested here and the canonical microcircuit (CMC) described by Bastos et al. \cite{bastos2012canonical}. When comparing the structure for a single predictive coding layer $l$, there is a clear correspondence between cause states, hidden states and associated prediction errors in biological and artificial networks. In the context of cortical function, a layer in CMC sends forward prediction errors to higher cortical areas and backward predictions towards lower areas. While this structure is visible in GPC, it lacks the explicit differentiation of states into superficial pyramidal cells and deep pyramidal cells. In the CMC, this differentiation can be connected to different sampling frequencies associated with incoming and outgoing information \cite{bastos2012canonical}.

The comparison with canonical circuitry and concrete neurophysiological evidence provides a useful starting point for more in depth analysis of the biological plausibility of computations in deep predictive coding networks (and for DNNs in general). One striking aspect in CMC is that feedforward information typically is propagated via \emph{linear} connections, while backward predictions feature \emph{non-linear} connectivity. GPC and PredProp consider this differentiation by including DNNs into the backward weights, while keeping precision and transition pathways linear.\footnote{PredProp, however, theoretically allows to to include non-linearities and complex autoregressive DNN models into the remaining connections between state variables.} Finally, comparing with the biological interpretation in CMC allows to address inhibitory and excitatory jointly with associated sampling frequencies in deep predictive coding explicitly. 

The hypothesized canonical circuitry provides an important bridge between analysis of predictive coding from a functional and computational level of analysis to concrete biological implementation. Similarly, GPC provides a concrete implementation of generalised predictive coding in the context of deep neural networks. 

One interesting line of research could address more exact modelling of the CMC and its implications on performance in machine learning tasks. Another, quite elegant way of characterising aspects of functional modularity in concrete implementations is to define functional modules in terms of statistical independence, i.e. by their Markov Blankets. We will use this approach in the following section in order to analyse differences between probabilistic graphical models describing generalized predictive coding and its concrete DNN-based implementation in GPC.

\subsection{From neurons to graphical models via Markov blankets}

\begin{figure}[H]
\includegraphics[width=\textwidth]{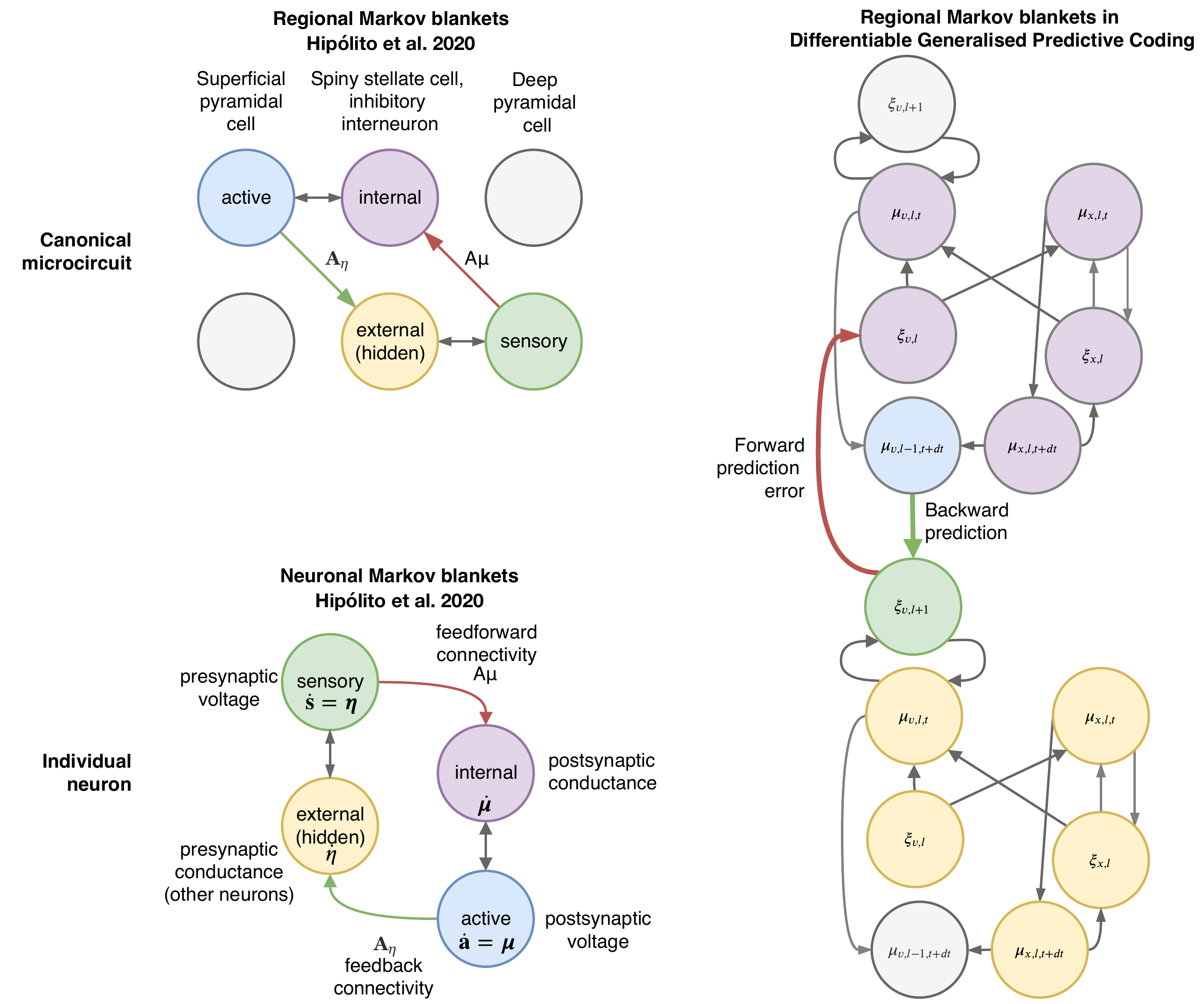}
\caption{Left: Neuronal Markov blankets describe that separate individual neurons and their surrounding. Regional Markov Blankets characterize the effective connectivity between cortical layers based on canonical microcircuits for predictive coding. Right: Similarly, regional Markov blankets can be mapped to the connectivity between predictive coding layers in Differentiable Generalised Predictive Coding.}
\label{fig:markov_blankets}
\end{figure}

A relatively recent line of research addresses the formalisation of dynamical coupling in biological systems across (spatial) scales via Markov Blankets \cite{friston2019free, ramstead2018answering, palacios2020markov, kirchhoff2018markov, pearl1998graphical}. Markov blankets describe a partitioning of complex dynamical systems in terms of statistical boundaries. When applied to characterising functional modules in biological brains, this entails partitions appearing at multiple scales, such as at the level of individual neurons, at the level of canonical micro-circuitry or at the level of brain regions and larger networks \cite{hipolito2021markov}. Markov blankets are an effective way to characterise statistical modularity in complex systems and can directly be mapped to the effective connectivity described by the model suggested here. Figure \ref{fig:markov_blankets} shows a simplified overview of Markov blankets at the neuronal and micro-circuit level, next to a mapping to the connectivity structure of GPC. While we characterize Markov Blankets at the regional level here and focus on relatively shallow GPC models, future work might characterize Markov Blankets appearing at larger spatio-temporal scales in more complex implementations.

Crucially, Markov Blankets divide the states of a systems into  external, sensory, active and internal states \cite{hipolito2021markov}. Of these, sensory and active states can be summarized into blanket states, which shield internal and external partitions. Internal states are updated based on internal and blanket states, but are independent of external states. Similarly, active states do not depend upon external states. As shown in Figure \ref{fig:markov_blankets} the level of canonical micro-circuits, one can separate neighboring layers in GPC into internal and external states, where information is propagated exclusively through active and sensory state, the state predictions and prediction errors respectively. An important aspect Markov blankets is that the described modularity is characterised exclusively by statistical dependencies and is not directly dependent on the physical structure of the systems. 

This differentiation between a system's structure in terms of Markov Blankets and the implementation can be applied in straightforward fashion to artificial systems modelled to mimic particular biological function or are tailored to a particular graphical model. The computations in the model presented here is based on the graphical model described by Generalised Predictive Coding. As visible in Figure \ref{fig:diagram} the implementation of the graph described by Generalised Predictive Coding seems to require exact connectivity to maintain a valid structure. For example, interconnecting variables within dynamical and hierarchical pathways intuitively seems to change the implemented graphical model. Future work, however, could relax the assumption of such strict mapping between connectivity in the graphical model and the implementation of GPC by allowing more flexible (i.e. random up to a certain degree) connectivity between variables. We can then characterize particular nodes in the model via their Markov blankets instead of a one-to-one mapping from functional to implementational level.

\section{Planning with Generalised Predictive Coding}

The hierarchical-dynamical transition model learned with GPC can be used for planning task in relatively straightforward matter. While arbitrarily complex algorithms could be employed on top of the learned model-based predictive model. Here, we focus on planning similar to recent formulations of Active Inference agents that optimise the \emph{Expected} Free Energy of with respect to policies over discrete actions \cite{ueltzhoffer2018deep, ccatal2020learning, van2020deep, fountas2020deep}. Policies $\pi$ are sequences of actions $a_t$ at discrete time step $t$. In many reinforcement learning algorithms, one would optimise on sequences of actions and observed rewards $r_t$:

$\tau=(a_0, r_0, a_1, r_1, ... , a_T, r_T)$ up to a planning horizon T. Typically, reward based methods require additional regularization that enables exploration in order to improve generalization. Such exploration often refers to testing new policies that might not be connected to (immediate) optimal reward and is typically modelled by including a random sampling process or some form of expected information gain.  

The approach in active inference is slightly different and does not directly operate on a known reward function.\footnote{External reward, however, can be integrated by defining a prior preference on maximising observed reward.} Active inference models optimises the Expected Free Energy (EFE), i.e. the Free Energy of expected observations for planned policies. Instead of resolving to external rewards, we can define actions over adaptive time intervals $dt_a$ and the associated Expected Free Energy:

\begin{equation}
\begin{aligned}
\tau_{dt_1} &= (a_1, EFE_{a_1}) \\
\tau_{dt_2} &= (a_2, EFE_{a_2}) \\
... &= ... \\
\tau_{dt_N} &= (a_N, EFE_{a_N}) \\
\end{aligned}
\end{equation}

in N parallel policies with discrete actions $a_n$ over planning horizons $dt_n$ with adaptive size. The adaptive sampling intervals in GPC implies that the temporal\footnote{Alternatively, the spatial interval size, when planning is done with respect to spatial dimensions. In hidden layers, the adaptive sampling size refers to the predicted motion of causes represented at the respective lower layer.} interval size for each action $a_t$ depends on the variational Free Energy of the previously inferred state. 

We can interpret this EFE objective as a sort of performance measure and simply optimise it with Stochastic Gradient Descent, similar to existing work on active inference interpreted as a policy gradient method \cite{millidge2020deep}.

\section{Nested hierarchies and perceptual actions}

\begin{figure}[H]
\begin{subfigure}{.8\textwidth}
  \centering
  \includegraphics[width=.8\linewidth]{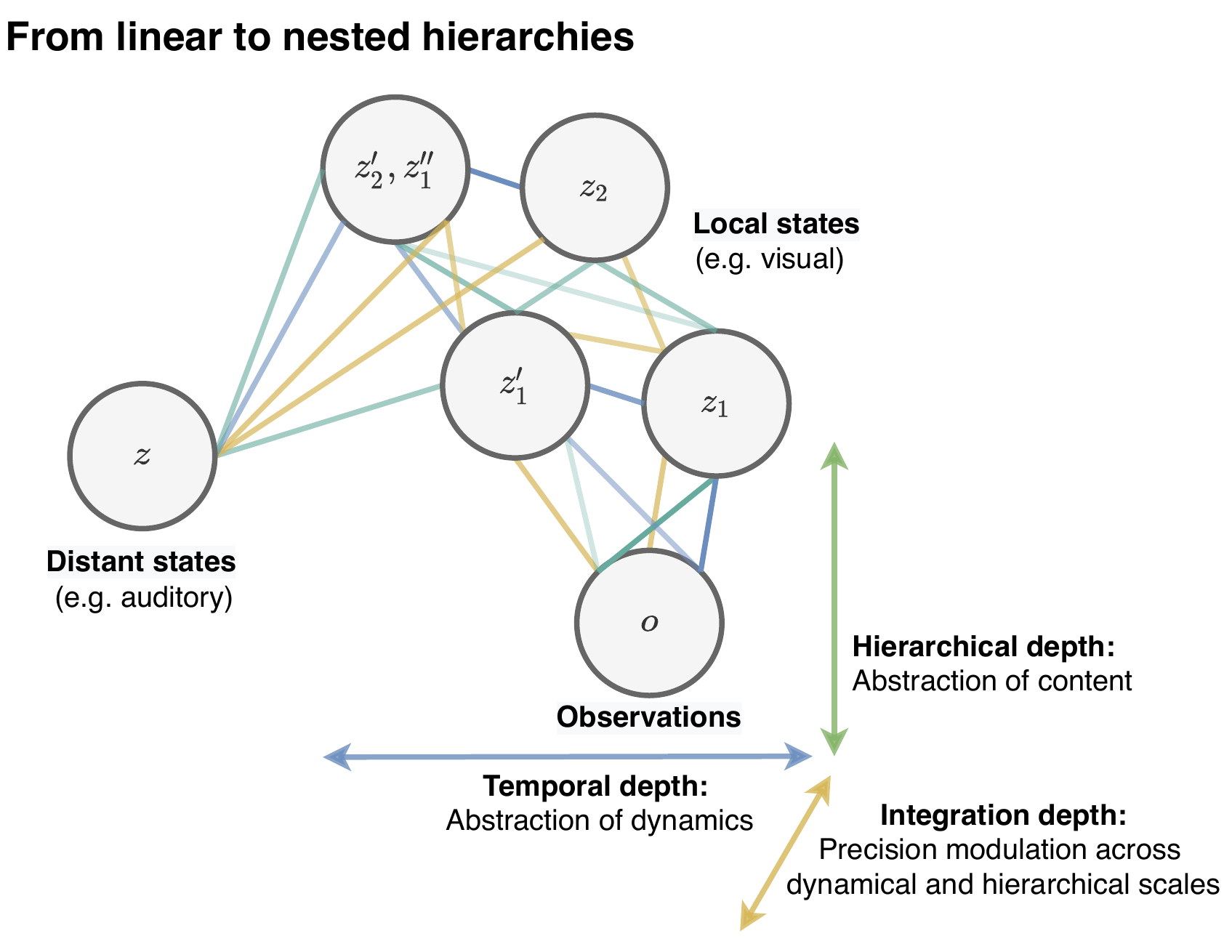}  
  \caption{A predictive coding network with four local and one distant nodes with nested hierarchical-dynamical structure.}
  \label{fig:sub-first}
\end{subfigure}
\begin{subfigure}{.8\textwidth}
  \centering
  \includegraphics[width=.8\linewidth]{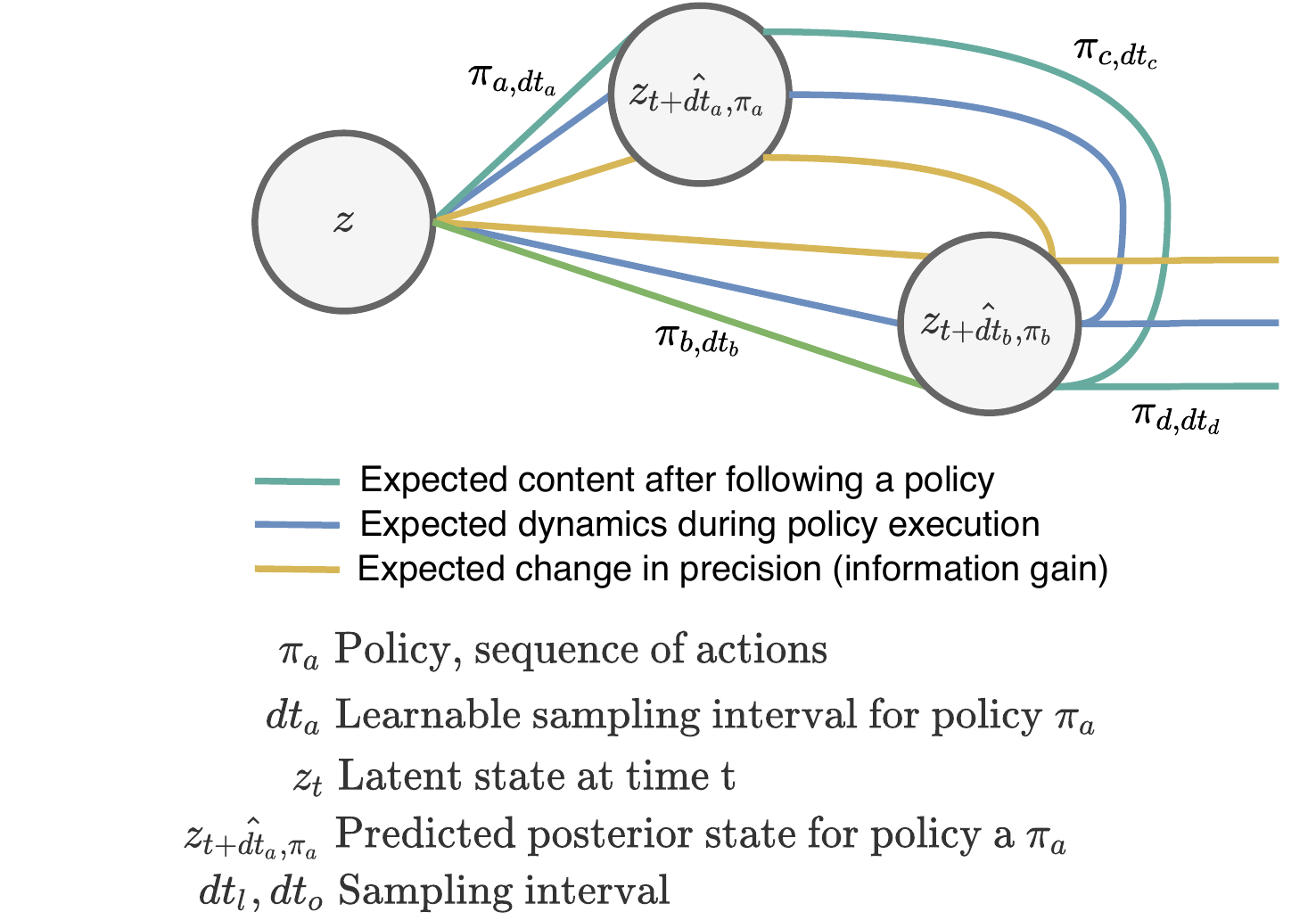}  
  \caption{Example trajectory with redundancy that can be reduced during perceptual model reduction in mental simulation.}
  \label{fig:sub-second}
\end{subfigure}
\caption{Perception and planning in generalized predictive coding networks with learned connectivity. }
\label{fig:fig}
\end{figure}

\printbibliography

\end{document}